# SinSpell: A Comprehensive Spelling Checker for Sinhala

Upuli Liyanapathirana
*University of Moratuwa*
upulilochana.16@cse.mrt.ac.lk

Kaumini Gunasinghe
*University of Moratuwa*
kauminikg.16@cse.mrt.ac.lk

Gihan Dias
*University of Moratuwa*
gihan@uom.lk

***Abstract*— We have built *SinSpell,* a comprehensive spelling checker for the Sinhala language which is spoken by over 16 million people, mainly in Sri Lanka. However, until recently, Sinhala had no spelling checker with acceptable coverage. Sinspell is still the only open source Sinhala spelling checker.**

**SinSpell identifies possible spelling errors and suggests corrections. It also contains a module which auto-corrects evident errors.**

**To maintain accuracy, SinSpell was designed as a rule-based system based on *Hunspell.* A set of words was compiled from several sources and verified. These were divided into morphological classes, and the valid roots, suffixes and prefixes for each class were identified, together with lists of irregular words and exceptions.**

**The errors in a corpus of Sinhala documents were analysed and commonly misspelled words and types of common errors were identified. We found that the most common errors were in vowel length and similar sounding letters. Errors due to incorrect typing and encoding were also found. This analysis was used to develop the suggestion generator and auto-corrector.**

## I. Introduction

Sinhala [1] is an official language of Sri Lanka and is spoken by over sixteen million people. It is a morphologically rich Indo-Aryan [2] language.

Spelling checking is a process of detecting misspelled words in text and providing suggestions for correction. Errors may also be automatically corrected. Spelling checkers are a basic part of word processing, text and chat applications. They are also used in automatic speech recognition (ASR) systems, text to speech (TTS) systems, machine translation systems, optical character recognition (OCR) systems, etc.

Errors may be due to incorrect usage, typing errors, encoding errors, OCR errors, ASR errors, etc. A spelling checker should build a model of such errors [3] and identify the probable correct form for each error. Corrections comprise one or more of insertion, deletion or substitution [5]. Incorrect word breaking (i.e. incorrectly including or omitting a space) may also be considered a form of spelling error.

This research was supported by the Accelerating Higher Education Expansionand Development (AHEAD) Operation of the Ministry of Education funded by the World Bank.

There are two main types of spelling errors [4] - non-word errors and real-word errors. A non-word error occurs when a word is neither in a dictionary, nor is an inflected form of a dictionary word. A real word error occurs when a word is in a dictionary but incorrect in the context, e.g.,

"I want to eat a peace of cake"

The correct word should be 'piece'.

The three main approaches used for spelling checking are (i) dictionary-based, (ii) data-driven and (iii) machine learning-based.

The dictionary-based approach [6] is the oldest spelling checking technique. Such a spelling checker contains a set of root words and a set of related inflections. Each input word is checked to see if it can be formed from a root word and an inflection.

The data-driven approach [5] compiles a list of correct words appearing in a corpus. It may be combined with a dictionary to capture uncommon words.

Machine learning [7] may be used to identify valid patterns, both within a word, and in the context in which a word is used. It may be used to detect *real-word* errors which are difficult to detect using the previous methods.

Although many spelling checkers exist for English and other widely used languages, very few spelling checkers are available for Sinhala. These include Subasa [13] and Microsoft Word. At the time of commencement of this project, neither of these were very accurate. However, the performance of MS Word has improved during the duration of this project.

In Section II, we review the current state of spelling checking in Sinhala and other languages. Section III describes the work we have done, and our results are presented in Section IV. Section V presents our conclusions.

## II. Literature Review

Computer-based systems for checking spelling mistakes have existed since 1961 [6]. Les Earnest is considered to have built the first spelling checker. Several spelling checking libraries like spellsoftware, ispell1, aspell2, myspell, Hunspell and other derivatives [4] may be used to build spelling checkers .

Hunspell [19] is a popular spelling checking engine which is based on myspell. Though originally developed for Hungarian, it supports many languages, and has good support for morphologically-rich languages. It is the default spelling checker in OpenOffice, Libre Office, Mozilla

Firefox & Thunderbird, Google Chrome, Mac OS X and Opera.

Spelling checkers have been built for Indic languages [7] such as Punjabi [8], Telugu [10], Malayalam [10],[11] and Tamil [12]. These systems use several approaches, such as dictionary-based [7], Morphology-based [8], data-driven [5],[13],[14] and deep learning based [9], [10], [15], [16], [17].

When considering the Sinhala language, we find only a few spelling checkers and their performance is similar to other Indic languages spelling checkers.

MS Word has an extension for Sinhala spelling checking. At the commencement of this research, it did not cover many words and error suggestions were also poor. However, it is now much improved.

Libre Office generally uses Hunspell spelling checkers as extensions. The Hunspell Sinhala dictionary created by Laknath [18] is a bit old and not compatible with current LibreOffice [20].

The "Subasa" spelling checker [5] was developed by Wasala in 2012 and can be considered an adequate tool. The author focused on detecting and correcting non-word errors under three factors: pronunciation and orthography of aspirated and unaspirated consonants, retroflex and dental letter confusion, and retroflex and palatal sibilants. The spelling checker uses an algorithm based on n-gram statistics computed from the UCSC Sinhala Corpus. It creates a unique word list and then a set of permutations of these words. The best suggestion module uses n-gram statistics to provide suggestions. Subasa also includes an auto-corrector feature.

Initially, Subasa only handled substitution errors and not addition, interchange, or deletion errors. In the second version of Subasa [13], the authors improved the approach to fill that gap using minimum edit distance techniques. Subasa only handles non-word errors and not real-word errors and context-based errors. It is available on-line at [21].

Buddhinie proposed a spelling checker in 2018 [14]. In this method permutation generation is based on similar structure, similar sounding letters and minimum edit distance method. Suggestions are given based on context.

## III. METHODOLOGY

SinSpell comprises two components as shown in Fig. 1:

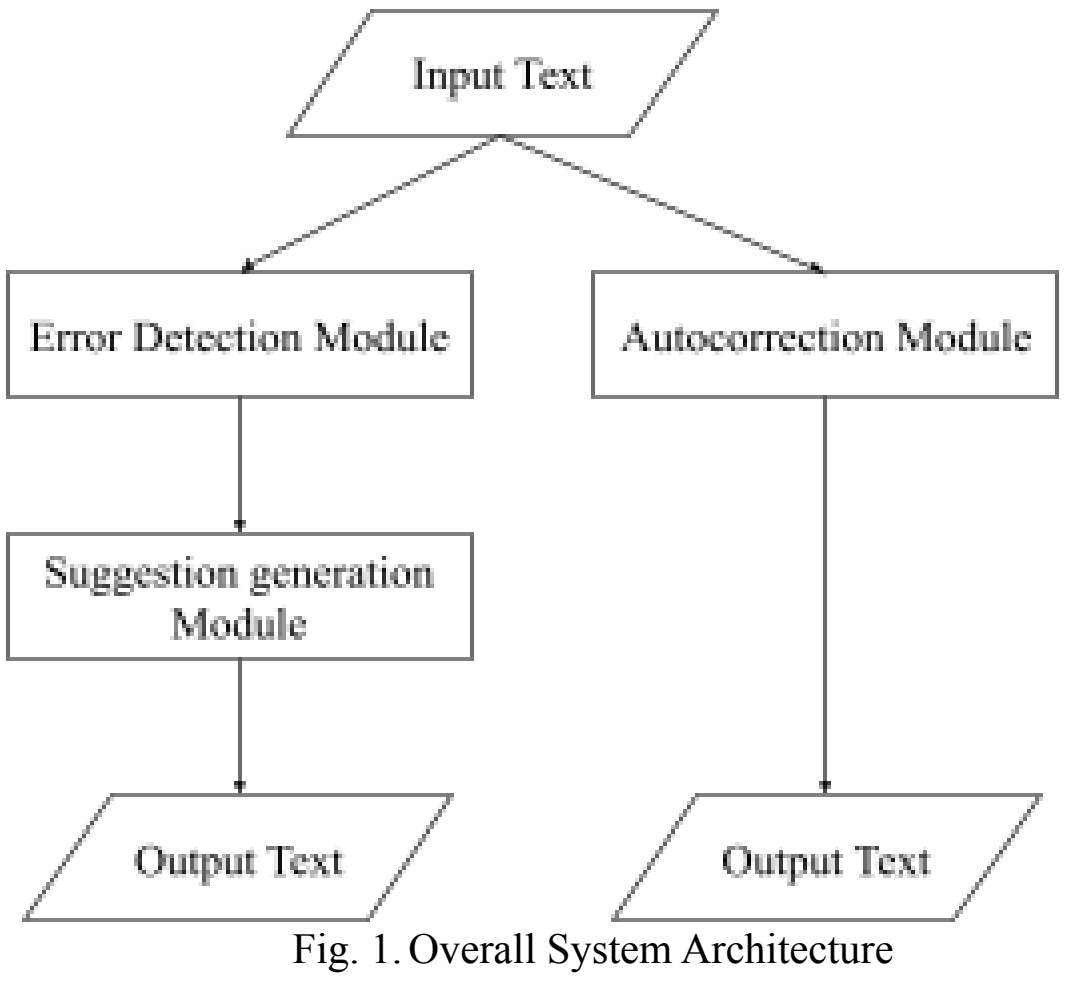

Fig. 1. Overall System Architecture

1. Spelling error detector and suggestion generator – A rule-based system using the *Hunspell* engine.
2. Auto spelling corrector – a finite-state transducer-based system.

These are described below.

### *A. Dictionary Creation*

Hunspell uses a dictionary file and affix file to identify correct words. We created custom dictionaries and affix files for Sinhala by obtaining data from several sources, including the Language Technology Research Laboratory, UCSC [23], the Concise Sinhala Dictionary [24], government documents and other sources.

Since the Sinhala morphological analyser - SinMorphy [25] - also required similar data, collection of data for the two projects was combined. We concentrated on adjectives and adverbs, while the SinMorphy team handled nouns, verbs and particles.

#### *a) Word lists*

To compile the lists of adjectives and adverbs, we compiled a list of unique words from several sources such as official documents, news reports, web crawlers, etc. Each word was tagged using a POS tagger [26]. We then generated separate word lists for each tag. In addition to identifying words for our project, these lists may be used by other Sinhala NLP projects. However, in their original form, they were not very accurate and needed further processing.

In addition to the adjectives and adverbs described below, we also extracted pronouns and particles from the POS tagged lists and added them to our dictionary after manual cleaning.

We also added some proper nouns such as city names to the dictionary..

#### *b) Adjectives and Adverbs*

Adjectives and adverbs were collected from the POS-tagged data, the Concise Dictionary, and other sources. Thereafter the word stems were identified, together with the set of possible suffixes for each adjective and adverb. A table was populated with the words in our list, and then manually expanded to include each valid suffix for each word. An extract is shown in Table 1.

Table 1: Examples of Adjective Suffixes

| Word | ම | ව | වම | වට | ත් | තම | වත් |
|---|---|---|---|---|---|---|---|
| විශාල | 1 | 1 | | 1 | | 1 | |
| වැඩි | 1 | 1 | | 1 | 1 | 1 | |
| අවශ්‍ය | 1 | 1 | | | 1 | 1 | |
| අබිලි | 1 | 1 | | 1 | | | 1 |
| තාවකාලික | 1 | 1 | 1 | | | | |
| සාමාන්‍ය | 1 | 1 | 1 | 1 | | | 1 |

The adjective and adverb data were shared with the SinMorphy project.

*c) Words from the morphological analyser*

We obtained noun and verb lexicons from SinMorphy. Lexicon files were in the *.lexc* format which consists of lists of word stems in each class, the suffixes which may be added to words of each class, and syntactical tags. Sinhala nouns were categorized into 26 classes. We wrote Python scripts to extract the stems and suffixes for each noun type from the lexicon files and store them in Hunspell dictionary and affix files. Following the same methodology, we extracted verb data from the lexc files, However, as verbs have a more complex representation, we only wrote a script to derive the past stem from the present stem and converted the rest of the data manually. An example noun lexicon file is shown in Fig. 2 below.

```
!!!nouns.lexc!!!

Multichar_Symbols N +RT +SG +DF +NOM +CJ +NOM +ACC
LEXICON Root
NounUncountableConsEndStem;

LEXICON NounUncountableConsEndStem


අප්‍රේල්     NounUncountableConsEnd;
අලුමීරිස්   NounUncountableConsEnd;
අයිස්      NounUncountableConsEnd;
අර්ගස්     NounUncountableConsEnd;
අසමොද්ගම්NounUncountableConsEnd;
අළුකෙහෙල්          NounUncountableConsEnd;
ඇපල්      NounUncountableConsEnd;
ඉළුක්      NounUncountableConsEnd;
ඊමේල්     NounUncountableConsEnd;
උලුවස්     NounUncountableConsEnd;
ඒඩ්ස්      NounUncountableConsEnd;


LEXICON NounUncountableConsEnd

+N+RT+UN+ACC:්#;
+N+RT+UN+NOM:්#;
+N+RT+UN+ACC+CJ:ුත් #;
+N+RT+UN+NOM+CJ:ුත් #;
+N+RT+UN+ACC+FN:ුයි #;
+N+RT+UN+NOM+FN:ුයි #;
+N+RT+UN+INT:්වලින් #;
+N+RT+UN+ABL:්වලින් #
+N+RT+UN+INT+FN:්වලිනි #;
+N+RT+UN+ABL+FN:්වලිනි #
+N+RT+UN+INT+CJ:්වලිනුත් #;
+N+RT+UN+ABL+CJ:්වලිනුත් #;
```

Fig. 2.Sample Noun Lexicon

Although many Sinhala words (e.g. nouns, verbs) take prefixes (උපසර්ග), the applicable prefixes do not fall neatly into classes. Therefore, we did not, in general, combine words with prefixes, but listed each prefixed word separately. However, the prefix නො- (no-), used to form the negative form of verbs, was included. Note that if a compound verb is written as a single word, the prefix is applied to the base verb, and becomes an *infix* of the compound word, e.g. ළංකරනවා → ළංනොකරනවා.

Table 2 shows some example root words, suffixes and prefixes.

TABLE 2. HUNSPELL SUFFIXES AND PREFIXES

| Root word | SFX-suffix words | PFX-<br>prefix words |
|---|---|---|
| අම්ම | අම්මා<br>අම්මට<br>අම්මගෙන් | |
| මිහිර | මිහිරි | අමිහිර |
| අංගනා | අංගනාව<br>අංගනාවෝ | |
| පිරිසිදු | පිරිසිදුව | අපිරිසිදු |

*B. Error Detection*

Error detection is performed by Hunspell using the dictionary and affix files constructed by us. Only non-word errors are detected.

*C. Error Analysis*

We had access to a large number of government documents such as annual reports, circulars, official letters, gazettes etc. In addition to the original version, we had versions corrected by a professional in Sinhala language.

We used this data to identify common errors in Sinhala documents. The documents were first aligned using the Hunalign tool[22] to bring sentences into alignment. The aligned documents were processed by a set of Python scripts to identify the changes made by the correctors. Using these tools, we obtained a set of erroneous strings with their respective corrected strings, as shown in Fig. 3.

- දිගටම + දිගට + ම
- ඒකම + ඒක + ම
- ස්ථාන + ස්ථානවල
- බැවිසහ + බැවින් + සහ
- පරිදී + පරිදි
- දරණ + දරන
- පරීපාටික + පරිපාටික
- සිදුවන + සිදු + වන
- අපේක්ෂක්ෂා + අපේක්ෂා
- කිර්මට + කිරීමට
- වෂයේ + වර්ෂයේ
- ආරක + ආරම්භක
- යටත්ව + යටත් + ව
- කටයුතු - වලට + කටයුතුවලට
- ඉල්ලීමව + ඉල්ලීම් + ද
- වූද + වූ + ද
- සානුකම්පිතව + සානුකම්පිත + ව

Fig. 3.Common errors in documents

In this figure, "-" indicates the incorrect strings, and "+", the corrected strings.

The error list was analysed to identify the following types of errors: *deletion, insertion, substitution* and *word separation*, as shown below. As vowel length errors were very common, they were listed as a separate type.

- Deletion e.g.: -සවලදායිත්වය +වලදායිත්වය, -මූලාශ්‍රාය +මූලාශ්‍රය, -පරිශ්‍රායක් +පරිශ්‍රයක්
- Insertion e.g.: -දැක් +දැක්වූ, -පිළිඳ +පිළිබඳ,
- Vowel length e.g.: -කිරිම +කිරීම, -අලුතින් +අලුතින්
- Substitution eg: -දැණුමින් +දැනුමින්, -තුලින් + තුළින්, -නොහැරීම +නොහැර්ම
- Word separation eg: -ලේකම් -වරයෙක් +ලේකම්වරයෙක්, -කාර්යමණ්ඩල +කාර්යය +මණ්ඩල

*D. Suggestion Generation*

Hunspell contains a built-in suggestion generator based on edit distance.We augmented this engine to better handle common spelling errors in Sinhala, as identified above. For example, vowel modifiers (පිලි) such as ් , ී, ි ,ා, ැ, ්‍ර ,ැ, ෑ, are often used incorrectly. For example, the word ඨට may be misspelled as ඪට or රූට. We included these as TRY characters in Hunspell to prioritise suggestions for these errors.

TRY ් ා ෘ ු ැ ූ ෑ ි ී ෙ

Fig 4. Single-character suggestions list

We also added replacement rules for murthaja (මූර්ධජ) and mahaprana (මහාප්‍රාන) letters, ස and ශ, rakaransaya (රකාරාංශය), etc., so if a misspelled word differs from a correct word only by one of the pairs in Table 3, Hunspell prioritises that suggestion.

TABLE 3. REPLACEMENT RULES (BIDIRECTIONAL)

| | | | |
|---|---|---|---|
| න | ණ | ර | ්‍ර |
| ල | ළ | ට | ඨ |
| ස | ෂ | ක | ඛ |
| ස | ශ | ඩ | ඪ |
| ච | ඡ | ඊ | ඉ |
| බ | භ | ප | ඵ |
| ද | ධ | | |

*E. Auto-correction*

The objective of the auto correction module is to identify words which are definitely in error, and to correct them to the words the author should have used. However, neither of these objectives is fully achievable, as an author may use a term not known to the spelling system, or may have meant to use a term different from the defined correction.

Accordingly, we decided to limit the scope of the autocorrector to (i) only correct input words which are very unlikely to be correct in our domain (which is government documents) and (ii) the correction made is very likely what the author should have written.

Therefore, unlike the spelling checker which identifies correct words and marks all others as incorrect, the autocorrector only corrects a small number of specific words. We obtained these words from The Sinhala guide from the Dept. of Official Languages [27] and the errors identified by us. Each correction is defined as an *incorrect:correct* pair as shown in Fig. 5.

| | |
|---|---|
| ලබාගෙන:ලබා ගෙන | නියැඳි:නියැදි |
| සිදුවන:සිදු වන | නිරික්ෂණ:නිරීක්ෂණ |
| පුද්ගල බාවය:පුද්ගලභාවය | නිශ්කාශන:නිෂ්කාශන |
| නිෂප්‍රයෝජන:නිෂ්ප්‍රයෝජන | නියග:නියඟ |

Fig. 5. Sample autocorrector data

When text is input to the autocorrector, each LHS string is substituted by the corresponding RHS string. In addition to full-word errors, we also defined errors in prefixes and suffixes. Word separation errors are also handled. The autocorrector was implemented in *Foma* and *Lexc*.

*F. Libreoffice extension*

A LibreOffice extension was created according to the LibreOffice documentation. This extension can be easily downloaded by user and installed in LibreOffice application to enable Sinhala spelling checking.

*G. SinSpell Website*

SinSpell website is created mainly using JavaScript, NodeJS, HTML, Bootstrap. It uses the created custom dictionary and affix files to perform error detection and suggestion generation.

The Sinspell website is hosted at: https://nlp-tools.uom.lk/SinSpell/

## IV. EVALUATION AND DISCUSSION

We used sample data from 5 different sources: annual reports, official letters, newspaper articles, magazine articles and Wikipedia, as our test set. The test set contains 2374 words, of which 457 were error words and 1917 correct words. The test set was input to (i) Subasa.lk and (ii) MS Word current (2020) version as well as SinSpell. The output of the 3 sources were evaluated by a Sinhala professional. True positives (correct words not marked), false positives (incorrect words not marked), true negatives (incorrect words marked as incorrect) and false negatives (correct words marked as incorrect) were identified for each of the three systems.

-

### A. Correct word detection

Number of correct words:1917

TABLE 4. CORRECT WORD DETECTION EVALUATION

| spelling checker | True positives | False negatives | True positive % |
|---|---|---|---|
| Word | **1897** | 20 | **98.9** |
| Subasa | 1912 | 5 | 99.7 |
| Sinspell | 1853 | 64 | 96.6 |

### B. Incorrect word detection

Number of incorrect words:457

TABLE 5. INCORRECT WORD DETECTION EVALUATION

| spelling checker | True negatives | False Positives | True negative % |
|---|---|---|---|
| Word | 420 | 37 | 91.9 |
| Subasa | 107 | 350 | 24.0 |
| Sinspell | **449** | 8 | **98.2** |

All 3 systems correctly identified most correct words. However, MS Word's vocabulary appeared to be better than SinSpell's as it correctly identifies more words. Although Subasa identified more correct words, its performance in identifying incorrect words was very poor. Sinspell gave the best performance in identifying incorrect words and had the lowest number of false negatives.

To evaluate the suggestion generator, we compiled a list of 100 incorrect words. We then checked the suggestions given for each of these by the 3 systems. We used mean reciprocal rank (the average of the reciprocals of the rank of results) to evaluate the suggestions for the incorrect words.

TABLE 6. SUGGESTION GENERATION EVALUATION

| SpellChecker | No. of Misspelled Words | 1st suggest-ion accuracy % | MRR |
|---|---|---|---|
| Word | 100 | 61.0 | **0.754** |
| Subasa | 100 | 69.0 | 0.690 |
| SinSpell | 100 | **62.3** | 0.729 |

We see that SinSpell performs best for 1st suggestion accuracy, and is comparable with MS Word in mean reciprocal rank.

## V. CONCLUSION

When this project commenced in 2020, the performance of the only available Sinhala spelling checkers - MS Word and Subasa - were quite poor. By the time the project was completed, MS Word's performance had improved substantially. Despite this, SinSpell best identifies incorrect words and provides the best first suggestion. We plan to add more words to our vocabulary which will increase the error detection accuracy and reduce our false negative rate.

We also performed the first ever analysis of errors in Sinhala text, and identified common errors of the types *insertions, deletions. substitutions (including vowel length)* and *word separations.*

Our auto-spell-corrector, though having only a small set of rules, is able to correct many common errors. We plan to improve this too by identifying more errors and error patterns.

Our approach is entirely rule-based. Although we could have used a deep-learning based approach, we considered that as Sinhala spelling is quite regular, a rule-based approach with a comprehensive vocabulary would produce good results. Also, finding sufficient labelled (i.e., spelling corrected) data for supervised learning would have been difficult.

We did not cover real-word error correction. Deep learning techniques may be suitable for real-word error correction.

## REFERENCES

[1] "Sinhala Language", En.wikipedia.org, 2020. [Online]. Available: https://en.wikipedia.org/wiki/Sinhala_language.

[2] Erdosy, George, "The Indo-Aryans of ancient South Asia: Language, material culture and ethnicity". Berlin: Walter de Gruyter. ISBN 3-11-014447-6, 1995.

[3] Mohamed Zakaria Kurdi, "Natural Language Processing and Computational Linguistics: speech, morphology, and syntax", Volume 1. ISTE-Wiley. ISBN 978-1848218482, 2016.

[4] P. Sanket, "Language Models: spelling checking and Autocorrection", https://towardsdatascience.com/. [Online]. Available: https://towardsdatascience.com/language-models-spellchecking-and-autocorrection-dd10f739443c

[5] A. Wasala, R. Weerasinghe, and R. Pushpananda, "A Data-Driven Approach to Checking and Correcting Spelling Errors in Sinhala," February 2017, 2010.

[6] L. Earnest, "The first three spelling checkers," Sr. Res. Sci. Emeritus, no. 1961, p. 15, 2011.

[7] R. Kumar, M. Bala, and K. Sourabh, "A study of spelling checking techniques for Indian languages," JK Res. J. Math. Comput. Sci., vol. 1, no. 1, pp. 105–113, 2018.

[8] H. Kaur, G. Kaur, and M. Kaur, "Punjabi spelling checker Using Dictionary Clustering.Pdf," Int. J. Science, Eng. Technol. Res., vol. 4, no. 7, pp. 2369–2374, 2015.

[9] G. U. M. Rao, P. A. Kulkarni and C. Mala, "A Telugu Morphological Analyzer," International Telugu Internet Conference Proceedings , 28th - 30th September 2011.

[10] P. B, S. K.P. and M. Kumar, "A deep learning approach for Malayalam morphological analysis at character level", Procedia Computer Science, vol. 132, pp. 47-54, 2018.

[11] K. Manjusha, M. Anand Kumar, and Soman Kp, "Deep learning based spelling checker for Malayalam language," Journal of Intelligent and Fuzzy Systems · March 2018

[12] T. Dhanabalan, R. Parthasarathi, and T. V. Geetha, "Tamil spelling checker", In Proceedings of 6th Tamil Internet 2003 Conference, Chennai, Tamilnadu, India, 2003.

[13] E. Jayalatharachchi, A. Wasala and R. Weerasinghe, "Data-driven spelling checking: The synergy of two algorithms for spelling error detection and correction", International Conference on Advances in ICT for Emerging Regions (ICTer2012), 2012.

[14] L. G. B. Subhagya, L. Ranathunga, W. H. A. Nimasha, B. R. Jayawickrama, and K. L. Maha Liyanarachchi, "Data driven approach to Sinhala spelling checker and correction," 18th Int. Conf. Adv. ICT Emerg. Reg. ICTer 2018 - Proc., pp. 27–32, 2018, doi: 10.1109/ICTER.8615577.

[15] D. Han and B. Chang, “A maximum entropy approach to chinese spelling check,” Proc. Seventh SIGHAN Work. Chinese Lang. Process., no. October, pp. 74–78, 2013.

[16] P. Etoori, “Automatic Spelling Correction for Resource-Scarce Languages using Deep Learning,” pp. 146–152, 2018.

[17] G. Tyler, X. Mei, and P. Marc , “Personalized spelling checking using Neural Networks,” MA 02125-3393, USA

[18] B. Laknath, “Sinhala Hunspell Dictionary”, Github, [Online]. Available: https://github.com/laknath/Hunspell-Dictionary-Tools

[19] "Hunspell: About", Hunspell.github.io, 2021. [Online]. Available: http://hunspell.github.io/.

[20] "Dictionaries - Apache OpenOffice Wiki", Wiki.openoffice.org, 2021. [Online]. Available: https://wiki.openoffice.org/wiki/Dictionaries#Sinhala_.28Sri_Lanka.

[21] "සුබස", Speller.subasa.lk, 2020. [Online]. Available: http://speller.subasa.lk/spellerweb.py.

[22] "Hunalign", Cs.cmu.edu, 2021. [Online]. Available: https://www.cs.cmu.edu/afs/cs.cmu.edu/project/cmt-40/OldFiles/OldFiles/Nice/ehuber/GALE-2007-urdu/hunalign/readme.html. [Accessed: 19- Feb- 2021]

[23] R. Weerasinghe, D. Herath and V. Welgama, "Corpus-based Sinhala Lexicon," in Proceedings of the 7th Workshop on Asian Language Resources, ACL-IJCNLP 2009, Suntec, Singapore,, 2009.

[24] Sinhala Dictionary Compilation Institute, සංක්ෂිප්ත සිංහල ශබ්දකෝෂය *(Concise Sinhala Dictionary)*, Colombo, Dept. of Cultural Affairs, 2018.

[25] Kalindu Kumarasinghe, Gihan Dias and Indu Herath, “Sin-Morphy: A Morphological Analyzer for the Sinhala Language”, MerCon 2021.

[26] Fernando, S., Ranathunga, S., Jayasena, S., & Dias, G., “Comprehensive part-of-speech tag set and svm based pos tagger for Sinhala”. in Proceedings of the 6th Workshop on South and Southeast Asian Natural Language Processing (WSSANLP2016) (pp. 173-182) 2016.

[27] Dept. of Official Languages, රාජකාරි ලිපි සම්පාදනයට අත්වැලක් *(A Guide to Compiling Official Documents),* 2nd ed, Rajagiriya, 2019.